\documentclass[sigconf]{acmart}

\newcommand{\removelatexerror}{\let\@latex@error\@gobble}
\makeatother

\usepackage{booktabs}
\usepackage[english]{babel}
\usepackage[colorinlistoftodos]{todonotes}
\usepackage{comment}
\usepackage{hyperref}
\usepackage{enumitem}
\usepackage[]{algorithm2e}
\usepackage{algorithmic}
\usepackage{amsmath}
\usepackage{mathabx}
\usepackage{bm}

\acmArticle{4}

\copyrightyear{2018}
\acmYear{2018}
\setcopyright{acmcopyright}
\acmConference[RecSys '18]{Twelfth ACM Conference on Recommender Systems}{October 2--7, 2018}{Vancouver, BC, Canada}
\acmPrice{15.00}
\acmDOI{10.1145/3240323.3240386}
\acmISBN{978-1-4503-5901-6/18/10}

\begin{document}
\title{Large-scale Recommendation for Portfolio Optimization}

\newif\ifanonymous
\anonymousfalse

\newif\ifschedule
\schedulefalse

\newif\ifspace
\spacefalse

\newif\ifshort
\shorttrue

\ifanonymous
	\newcommand{\rakuten}{E-commerce Website Y }
	\newcommand{\gora}{[Anonymized] }
	\newcommand{\travel}{[Anonymized] }
\else
	\newcommand{\rakuten}{Rakuten Ichiba}
	\newcommand{\gora}{Rakuten GORA}
	\newcommand{\travel}{Rakuten Travel}

    \author{Robin M. E. Swezey}
    \affiliation{ 
           \institution{Rakuten Institute of Technology}
           \streetaddress{Rakuten Crimson House, 1-14-1 Tamagawa}
           \city{Setagaya-ku}
           \state{Tokyo}
           \country{Japan}
           \postcode{158-0094}
    }
    \email{rswezey@acm.org}
    
    \author{Bruno Charron}
    \affiliation{ 
           \institution{Rakuten Institute of Technology}
           \streetaddress{Rakuten Crimson House, 1-14-1 Tamagawa}
           \city{Setagaya-ku}
           \state{Tokyo}
           \country{Japan}
           \postcode{158-0094}
    }
    \email{bcharron@acm.org}
\fi

\newcommand{\transmat}{P_{CF}(S_{t+1}|S_t)}

\begin{abstract} 
Individual investors are now massively using online brokers to trade stocks with convenient interfaces and low fees, albeit losing the advice and personalization traditionally provided by full-service brokers.
We frame the problem faced by online brokers of replicating this level of service in a low-cost and automated manner for a very large number of users.
Because of the care required in recommending financial products, we focus on a risk-management approach tailored to each user's portfolio and risk profile.
We show that our hybrid approach, based on Modern Portfolio Theory and Collaborative Filtering, provides a sound and effective solution.
The method is applicable to stocks as well as other financial assets, and can be easily combined with various financial forecasting models.
We validate our proposal by comparing it with several baselines in a domain expert-based study.
\end{abstract}

\ccsdesc[500]{Information systems~Recommender systems}
\ccsdesc[300]{Theory of computation~Convex optimization}
\ccsdesc[300]{Applied computing~Economics}

\keywords{Recommender Systems; Collaborative Filtering; Modern Portfolio Theory}

\maketitle

\section{Introduction}
\label{sec:introduction}



Consider the case of an Internet service where individuals can buy, sell and hold financial instruments.
Unlike traditional brokers, online brokers do not usually provide personalized investment nor trading advice.
The simplest form of such advice is recommending to a user which stock they should buy next.
However, automated recommendations in online brokers are, to the best of our knowledge, either non-existent or non-personalized.  
One reason is regulations in the consumer financial industry; the other is the extra care needed in automating recommendations for high-stake products.
Recommendations thus need to follow a specific focus.

Our objective in this work is to recommend to users not only stocks they may like, but also satisfy a clearly explainable investment agenda.
Conversely, the objective is also not necessarily to optimize for returns: customers may likely prefer portfolios they are familiar with over those that provide better returns.
These specificities make it difficult to provide such recommendations without human supervision, which online brokers aim to reduce as much as possible in order to cut costs and offer the lowest fees.

In this paper, we propose an approach based on Modern Portfolio Theory (MPT) and Collaborative Filtering (CF).
The paper is structured as follows: in Section \ref{sec:related_work}, we situate this work in the context of existing research; section \ref{sec:foundation} is a primer on the required background in MPT and CF; in Section \ref{sec:approach}, we detail our approach and contributions.
We then detail our experiments in Section \ref{sec:experiments}, and conclude in Section \ref{sec:conclusion} with directions for future works.

\section{Related Work}
\label{sec:related_work}

Given the widespread access to market data, the purely quantitative nature of the problem and its financial appeal, there has been a large amount of literature devoted to the stock market.
Most efforts have been focused on identifying strategies for picking stocks~\cite{Geva2014,Gottschlich2014,Chou1996ASystem,Sun:2018:NSR:3234111.3234124,Wen2010AutomaticAlgorithm,Zhang2006StockRecommendations} or portfolios~\cite{Fasanghari2010,Nanda2010ClusteringManagement,Paranjape-Voditel2013AMining} which are likely to be profitable in the future.
Although these approaches are more akin to trading, they fall within the scope of non-personalized recommendation.
The literature on personalized recommendation of financial assets, on the other hand, is much sparser.
It can be argued that this is due to the lack of open data on traders behavior, as compared for example to research on personalized movie recommendation, where large-scale open data is available.

For the problem of recommending a single stock to the user, some personalized approaches based on technical analysis have been proposed~\cite{Chalidabhongse2006AModeling,Yoo2003AnAdvice} where the trading preferences of the user, in particular their response to technical signals, are learned through feedback on the recommendations.
Collaborative Filtering has also been used for personalization, in general combined with other recommendation logics, such as order book analysis~\cite{Yujun2016AnInflow}, content-based filtering~\cite{Taghavi2013Agent-basedSystem} or multiple-criteria decision analysis~\cite{Matsatsinis2009NewSelection}.

In the field of portfolio recommendation, Modern Portfolio Theory (MPT)~\cite{Markowitz1952PORTFOLIOSELECTION} and its treatment of the risk-return tradeoff serves as the theoretical basis for a significant part of the literature on non-personalized recommendation and for the industry at large.
However, personalized portfolio recommendations have rather been based on other approaches such as case-based reasoning~\cite{Musto2014} or psychological analysis~\cite{Garcia-Crespo2012}.
Recommendation personalized by the knowledge of the current portfolio of the user, which is the problem tackled in this paper, has been previously tackled in ref.~\cite{Zhao2015Risk-HedgedRecommendation} for the field of venture finance.
The authors propose several methods based on MPT to extend the current portfolio of users in an optimal way.
One of these methods recommends the best options to add an individual asset to the user's portfolio.
Our work expands on this approach by both combining it with Collaborative Filtering to improve the quality of the recommendations and scaling it to large number of users, which is arguably irrelevant in the field of venture finance but not in that of retail trading.

\section{Foundations}
\label{sec:foundation}

\subsection{Modern Portfolio Theory}
\label{ssec:mpt}

All recommendation approaches aim at presenting to the user the items that they would prefer.
Mathematically, this corresponds to maximizing a utility function $u$ specific to the user, satisfying $u(j) > u(j')$ if and only if the user would unambiguously prefer item $j$ over item $j'$.
Modern Portfolio Theory (MPT), introduced by Markowitz~\cite{Markowitz1952PORTFOLIOSELECTION}, proposes a form for the utility function of an investor over financial portfolios.
Given a fixed amount to invest and $n$ available assets $[1..n]$, a portfolio can be represented by a vector $\mathbf{w}$ whose components $w_j$, $j\in[1..n]$, are the proportion of the total amount invested on the asset $j$.
The MPT utility function has a single parameter, the risk aversion $\gamma$, to adjust the function to a particular user, and only takes into account the expectation ($\mathbb{E}$) and variance ($\operatorname{Var}$) of the future return $R_\mathbf{w}$ of the portfolio $\mathbf{w}$, modeled as a random variable given the uncertainty, in the form
\begin{equation}
u_\gamma(\mathbf{w}) = \mathbb{E}[R_\mathbf{w}] - \gamma \operatorname{Var}[R_\mathbf{w}].
\end{equation}

The distributions for $R_\mathbf{w}$ can be based on any financial theory, and the contributions of the paper are independent of a particular choice, as long as they are linear in $\mathbf{w}$.
For each asset $j\in[1..n]$, let $\mathbf{w}^j$ be the portfolio solely composed of asset $j$.
Then it follows from linearity that
\begin{equation}
\label{eq:mpt-util}
u_\gamma(\mathbf{w}) = \bm{\mu}^T \mathbf{w} - \gamma \mathbf{w}^T \Sigma \mathbf{w},
\end{equation}
where the vector $\bm{\mu}$ has components $\mu_j = \mathbb{E}[R_{\mathbf{w}^j}]$ and the matrix $\Sigma$ has components $\Sigma_{j,j'} = \operatorname{Cov}(R_{\mathbf{w}^j}, R_{\mathbf{w}^{j'}})$ for any assets $j, j'\in[1..n]$.

Portfolios $\mathbf{w}$ are traditionally visualized in the risk-return plane, with x-axis their risk $\sqrt{\operatorname{Var}[R_\mathbf{w}]}$ and y-axis their expected return $\mathbb{E}[R_\mathbf{w}]$.
As shown in (\ref{eq:mpt-util}), the utility function $u_\gamma(\mathbf{w})$ is concave in the portfolio weights $\mathbf{w}$ so that the optimal portfolio, maximizing the function at a given risk aversion $\gamma$, can be extracted numerically in an efficient way.
The set of optimal portfolios for $\gamma \in [0, \infty)$ form a curve in the risk-return plane, called the Efficient Frontier (EF), see Figure~\ref{fig:reco_EF} for an example.

For our particular implementation, we choose the random variables $R_\mathbf{w}$ to be distributed according to the sampling distribution of their historical realizations, each past realization of the returns being weighted by an exponential decreasing in its age.
This results for example in expectations $\mu_j$ at a given time to be exponentially-weighted moving averages of the historical returns of assets $j \in [1..n]$.
The exponential decay parameter and frequency of the returns can be tuned for the desired investment time horizon.

\subsection{Collaborative Filtering}
\label{ssec:cf}

The trade-off between risk and return modeled by MPT is not the only aspect considered by investors.
A major missing component is their familiarity with the available assets and we use Collaborative Filtering (CF) based on item-to-item conditional probabilities to model it.
Our rationale is: 

1. The number of users is orders of magnitude higher than the number of stocks and item-item matrices are dense.
Item turnover is low and cold start is not problematic.

2. Keeping user-level data to infer another user's recommendations (as in instance-based learners) can be legally challenging, as well as deterring for users.

3. Computation of CF does not bottleneck the time complexity achieved later for MPT recommendations (Section \ref{ssec:mptrec}).

Given $m$ users and $n$ items, for a time interval $T$ and an interaction function $f$ of interest, we define the $m\times n$ user-item matrix $U_{f,T}$ with components
\begin{equation}
(U_{f,T})_{ij} = f(i,j,T).
\end{equation}

Let $q_{i,j,t}$ be the market value of the position of user $i$ on stock $j$ on day $t$, which is obtained from daily snapshots of user portfolios.
We define the implicit feedback CF user-item matrix $R$ as $U_{f_R, T_R}$ where
\begin{equation}
f_R(i,j,T) = \begin{cases}
1 &{\text{if there is $t\in T$ s.t. }} q_{i,j,t} \neq 0 \\
0&{\text{otherwise}}
\end{cases},
\end{equation}
and the aggregate normalized user-portfolio matrix $W$ as $U_{f_W, T_W}$ where
\begin{equation}
f_W(i,j,T) = \frac{\sum_{t \in T} q_{i,j,t}}{\sum_{j' \in [1..n], t' \in T} q_{i,j',t'}}.
\end{equation}

Rows of the $R$ matrix capture, for user $i$, which stocks $j$ they held at least once during some interval $T_R$.
To build a dense enough item-item matrix, $T_R$ is chosen to span at least 6 months for $R$.
In the case of $W$, $T_W$ spans a shorter period such as the last day, week or month, adjustable depending on the investment horizon of the user.
Besides daily portfolio snapshots, transaction history can also be used to target traders instead of investors. (Note that aggregation from trading and holding data becomes similar as $T$ grows.)

The so-called \emph{co-count} item-item matrix is given by
\begin{equation}
\begin{aligned}
\widetilde{C} = R^T R 
\end{aligned}
\end{equation}
of computational time complexity $O(mn^2)$. The Markov item-item transition matrix $C$ defining $P(S_{j'}|S_j)$, with $(C)_{jj'}$ the probability of moving from stock $S_j$ to stock $S_{j'}$, is obtained by removing $\widetilde{C}$'s diagonal and normalizing along columns. We then leverage the fact that $W$ is equivalent to a collection of prior user state vectors $P(S_j|U_i)$, with row $\mathbf{w}^{(i)}$ the initial state of user $i$, to build the portfolio-based personalized CF recommendations $\widebar{Y}_{CF}$ using
\begin{equation}
\label{eq:usercf}
\begin{aligned}
\widebar{Y}_{CF} = WC
\end{aligned}
\end{equation}
Although the user portfolios are sound and convenient to use as prior probabilities in this context, no interpretation should be made on their reflecting relative \emph{preferences} from the user.
Finally, as for the computation of $\tilde{C}$, the computation of $\widebar{Y}_{CF}$ is also $O(mn^2)$.

\section{Recommending Stocks}
\label{sec:approach}
The following sections follow our overall pipeline:

1. Compute the EF of risk-return policies that could be achieved by any user, given a risk aversion range and all stocks, and map users to an estimate of their risk aversion.

2. For each (user, target stock) pair, compute the MPT utility of adding that individual stock to the user's portfolio.

3. Compute CF recommendations to restrict the available universe, then re-rank them with MPT scores from step 2.

\subsection{Estimating Users Risk Aversion}
\label{ssec:gamma}

As seen in Section~\ref{ssec:mpt}, the MPT utility function is personalized by a single parameter $\gamma$, the risk aversion of the user.
With access to the daily historical portfolios $\mathbf{w}_t$ of a user over a period $T$, we infer their risk aversion as follows.
On each day $t$, we define $\gamma_t$ as the risk aversion for which the optimal portfolio has the same risk as $\mathbf{w}_t$; visually the intersection of the efficient frontier with a vertical line passing through the portfolio of the day in the risk-return plane.
The rationale is that the user, accepting such a risk, would have picked that optimal portfolio given a complete knowledge and the absence of trading constraints.
The risk aversion of the user is defined as the geometric mean of the daily estimates $(\gamma_t)_{t\in T}$.

\begin{figure}[t]
 \centerline{\includegraphics[width=0.5\textwidth]{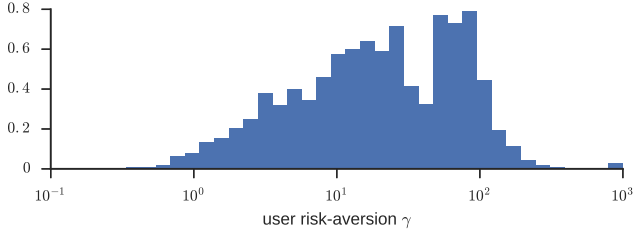}}
 \vspace{-.5cm}
\caption{\label{fig:hist_gamma}Distribution of the user risk aversion, with $T$ the month of April 2016.}
\end{figure}

Figure~\ref{fig:hist_gamma} shows the distribution of the risk aversion of the users in logarithmic scale, see Section~\ref{sec:experiments} for details on the data.
The risk aversions are distributed around the median $20.9$, with an accumulation around $100$, which corresponds to risk-averse users trading only low variance assets.

\subsection{Computing Efficiently the Utility of Individual Stock Recommendations}
\label{ssec:mptrec}

\begin{figure}[t]
 \centerline{\includegraphics[width=0.5\textwidth]{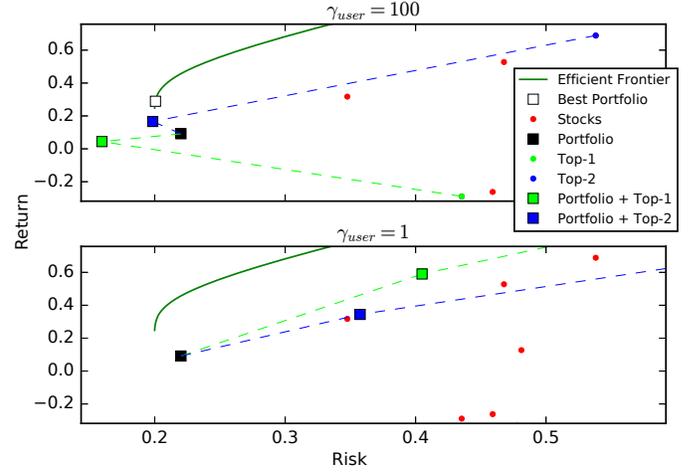}}
\caption{\label{fig:reco_EF}Movement of the user portfolio (black square) to new positions (green and blue squares) in the risk-return plane, once the user has bought either the top-1 (green dot) or top-2 (blue dot) recommended stock in suggested quantity. A high risk aversion $\gamma_{user}$ will favor low-risk over high-return recommendations. The white square shows the best achievable portfolio given all stocks and the specified $\gamma_{user}$.}
\end{figure}

As fees are expensive for users who rarely invest in more than one new stock per day, we target daily one-stock recommendations.
This means that, although the results are presented in a ranked list, (i) we expect users to invest in at most one new recommended stock per day and (ii) explain to them that each stock is picked based on its individual impact on the portfolio regardless of other recommendations.
Figure \ref{fig:reco_EF} gives a visual explanation of how recommendations are ranked on the risk-return plane.

We define the MPT recommendation matrix $\widebar{Y}_{MPT}$ whose $(i,j)$ coefficient is the user $i$'s portfolio utility once stock $j$ has been added to it with weight $w_r^{(i)}$, the average of the non-zero weights of the user's portfolio $\mathbf{w}^{(i)}$.
Formally,
\begin{equation}
\label{eq:y_mpt}
(\widebar{Y}_{MPT})_{ij} = u_{\gamma_i}((1 - w_r^{(i)}) \mathbf{w}^{(i)} + w_r^{(i)} \mathbf{e}_j),
\end{equation}
where the $n$-size unit vector $\mathbf{e}_j$ is zero except at index $j$.

The naive sequential way to compute such a matrix can be found in current literature such as \cite{Zhao2015Risk-HedgedRecommendation} (algorithm 3). Essentially, the quadratic utility function needs be computed for all $mn$ (user, target stock) pairs, leading to $O(mn^3)$ complexity.
For convenience and parallelism, one may further wish to derive the scoring matrix analytically, but doing so naively can result in \(O(m^2n^2)\) complexity.
However, if we leverage the expectation that recommendations are made on a one-stock basis, it becomes possible to compute the whole scoring matrix in closed form with \(O(mn^2)\) complexity as
\begin{equation}
\label{eq:y_mpt}
\begin{aligned}
W' &= ((1-\mathbf{w}_{r}) \mathbf{\hat{j}}^T) \circ W \\
\widebar{Y}_\mu &= W'\mathbf{\mu} \mathbf{\hat{j}}^T + \mathbf{w}_{r} \mathbf{\mu}^T \\
\widebar{Y}_\Sigma &= (W'\Sigma \circ W') \mathbf{\hat{j}} \mathbf{\hat{j}}^T
		 + (\mathbf{w}_{r} \circ \mathbf{w}_{r}) diag ( \Sigma )^T \\
		 &\quad+ 2W'\Sigma \circ (\mathbf{w}_{r} \mathbf{\hat{j}}^T) \\
\widebar{Y}_{MPT} &= \widebar{Y}_\mu - \bm{\gamma} \widebar{Y}_\Sigma
\end{aligned}
\end{equation}
with $\mathbf{w}_{r}$ the $m$-size vector of $w_r^{(i)}$'s, $\bm{\gamma}$ the $m\times m$ diagonal matrix of user risk aversions, $\mathbf{\hat{j}}$ the $n$-size all-ones vector, and $\circ$ the Hadamard / element-wise product.

The above analytical derivation is a desirable contribution of this work since (i) it scales linearly in users, (ii) is one less power to compute in stocks, and (iii) is a parallelizable vectorized form.

\subsection{Hybrid Recommendations}
\label{ssec:hybrid}

Because MPT recommendations do not account for the knowledge of users about stocks, and CF recommendations have weak utility, we propose a hybrid re-ranking approach:


i. Sort the available recommendations by CF score.

ii. Given a cutoff $k$, retain the top-k results of step (i).

iii. Sort the results of step (ii) by MPT scores.

The exact implementation is described in Algorithm \ref{algo:hybrid}.

\begin{algorithm}[ht]
 \textbf{Requires:} $\widebar{Y}_{CF}$, $\widebar{Y}_{MPT}$, $k$ \\
   \textbf{Initialize} $\widebar{Y}_H \gets \texttt{matrix}(m, n)$ \;
   $\widebar{Y}_H[:, :] \gets -\infty$ \\
   \For{$u = 1:m$}{
    	$\texttt{filtered} \gets \texttt{argsort}(\widebar{Y}_{CF})[1:k] $
      	$\widebar{Y}_H[u, \texttt{filtered}] \gets \widebar{Y}_{MPT}[\texttt{filtered}] $
 	}
 \caption{MPT-CF Hybrid Recommendations.}
 \label{algo:hybrid}
 \end{algorithm}

\section{Evaluation}
\label{sec:experiments}


The approach presented in this work aims to provide a completely automated alternative to the personalized recommendations of financial experts in traditional brokerage companies.
Rather than some rigorous quantitative measures, the customers' appraisal of the recommendations is usually based on the trust granted to such experts.
We have thus chosen to partner with a brokerage firm, to perform a domain expert-based study to validate our system.

For the purpose of this research we obtained the historical trading data of a few hundreds of thousands of users, for two years up to March 2017, on the stocks of 3,714
\ifanonymous
public companies in the partner firm's operating country (anonymized).
\else
companies listed on the Tokyo Stock Exchange and the Nagoya Stock Exchange in Japan.
\fi
This very large number of companies makes it virtually impossible for investors and even domain experts to take comprehensive decisions and highlights the need for a system such as the one presented in this work.

We picked 20 actual user portfolios spanning different sizes and industries and randomly associated them risk aversion parameters drawn from $\{1, 20, 100\}$.
For each, we generated the top-20 recommendations for the following 4 methods: Random (drawn randomly from the set of possible stocks), MPT (using the MPT scoring matrix $\widebar{Y}_{MPT}$, see~\eqref{eq:y_mpt}), UserCF (using the CF scoring matrix $\widebar{Y}_{CF}$, see~\eqref{eq:usercf}) and Hybrid (using the Hybrid scoring matrix $\widebar{Y}_{H}$, see Algorithm~\ref{algo:hybrid}).

Three experts working in the brokerage firm, with knowledge of the stocks market and without involvement in the current work, were selected as evaluators.
The evaluators where asked for each of the 20 portfolios and associated risk aversions to choose their preferred recommendations in two steps, each unlabeled and randomized:
first choosing amongst the top-5 recommendations of the 4 methods,
then choosing between the top-5, 6-to-10, 11-to-15 and 16-to-20 recommendations of the method selected in the first step.
This setup allowed us to compare not only the overall quality of the methods but also the ranking within each method.

To compare the 4 methods in the quality of their top-5 lists, we regard the first step of our evaluation as a randomized block design.
Blocking on the portfolio / risk aversion pairs, we create a 20 x 4 matrix containing for each pair and method the number of evaluators selecting that method as best on the pair.
Although the number of evaluators is relatively small, summing over their results should reduce the influence of individual specificities and the blocks can be considered almost independent.
Using the Friedman test, we obtain a p-value $p = 0.0014$ for the omnibus null hypothesis. 
As this result is significant (we use a 95\% level), we apply the Nemenyi method for post-hoc analysis and obtain p-values shown in Table~\ref{tab:pvalues} for pairwise comparisons of the methods.
Only the Hybrid method has significantly different performance than random recommendations, which confirms its value.
Table~\ref{tab:ratios} further shows that the Hybrid method dominates all baselines but as stated above the difference is only significant compared to the random recommendations.

\begin{table}[t!]
\centering
\begin{tabular}{l|c|c|c||c}
     & 3 eval. & 2 eval. & 1 eval. & Selections \\
\hline
Random & 0 & 1 & 4 & 6 (10\%)  \\
MPT & 0   & 0 & 7  & 7 (12\%) \\
UserCF & 1  & 4  & 10  & 21 (35\%) \\
Hybrid & 2   & 7  & 6 & 26 (43\%)
\end{tabular}
\caption{
\label{tab:ratios}
Number of portfolios for which each method was selected as best by 1, 2 or 3 evaluators and total number of such selections per method.}
\end{table}

\begin{table}[t]
\centering
\begin{tabular}{l|c|c|c}
     & MPT & UserCF & Hybrid \\
\hline
Random & 0.995 & 0.079 & \textbf{0.030}  \\
MPT & & 0.140 & 0.058 \\
UserCF & &  & 0.983
\end{tabular}
\caption{
\label{tab:pvalues}
Result of the post-hoc analysis for comparison of the 4 methods on their top-5 lists.
}
\end{table}

\begin{table}[t!]
\centering
\begin{tabular}{l|c|c|c|c}
 & (Intercept) & MPT & UserCF & Hybrid \\
\hline
Estimate & 0.69 & 0.22 & -0.40 & 0.31 \\
P-value & 0.42 & 0.85 & 0.68 & 0.75
\end{tabular}
\caption{
\label{tab:coeff}
Results of the ranking quality analysis.
}
\end{table}

To analyze the ranking quality, we use the logit regression $keep \sim MPT + UserCF + Hybrid$ where $keep$ is the binary response describing whether the evaluator keeps in step 2 the list accepted in step 1 and $MPT$, etc. are binary independent variables describing whether the associated method was chosen in step 1.
Coefficient estimates and associated p-values are shown in Table~\ref{tab:coeff}.
We can see with the positive intercept an overall tendency for evaluators to keep the list irrespective of the model.
The estimates for $MPT$ and $Hybrid$ are positive, hinting at a good quality of the ranking, and close as expected given that the hybrid recommendations are ranked using MPT after a filtering step.
The estimate for $UserCF$ on the other hand is negative suggesting that while doing a good job at picking good recommendations among the 3,714 companies, its ranking within the top-20 may not be very accurate.
Nonetheless, due to the small size of the data, none of these qualitative observations are statistically significant at a remotely reasonable level.

\section{Conclusion and Future Works}
\label{sec:conclusion}

In this work, we framed the problem of making personalized recommendations for users of an online broker, a context for which research remains scarce. 
We proposed a new and scalable approach tailored to users’ estimated risk profiles and affinities, combining Modern Portfolio Optimization and Collaborative Filtering in a hybrid algorithm.
We showed that the approach is sound, achieves attractive complexity, and we validated its value compared to multiple baselines in an experiment based on domain experts.

The method, being untied to the way expected returns and risks are computed, can be plugged to any type of forecast that provides those two types of data, so long as user's holding or transaction data is available. 
Likewise, the type of financial instrument is exchangeable, so long as recommendations can be made on a one-instrument basis.

\iftrue
\else
\section*{Acknowledgements}
We thank Young-joo Chung, Weichen Cai, Martin Rezk, Yu Hirate, and all contributors to this research for their support.
\fi

\bibliographystyle{ACM-Reference-Format}
\bibliography{sigproc.bib}

\end{document}